\pdfoutput=1

\documentclass[11pt]{article}

\usepackage[]{EMNLP2022}

\usepackage{times}
\usepackage{graphicx}
\usepackage{latexsym}
\usepackage{enumitem}
\usepackage{amsfonts} 
\usepackage{amssymb}

\usepackage[T1]{fontenc}

\usepackage[utf8]{inputenc}

\usepackage{microtype}

\usepackage{inconsolata}

\title{Conformal Predictor for Improving Zero-shot Text Classification Efficiency}

\author{Prafulla Kumar Choubey$^1$ \quad Yu Bai$^1$ \quad  Chien-Sheng Wu$^1$ \\
 {\bf\stepcounter{footnote}Wenhao Liu$^{2}$~\thanks{{ } work was done at Salesforce AI Research.} \quad Nazneen Rajani$^{3}$~\footnotemark[2]} \\
$^1$Salesforce AI Research, $^2$Faire.com, $^3$Hugging Face \\
\texttt{\{pchoubey, yu.bai, wu.jason\}@salesforce.com} \\
\texttt{wenhao@faire.com, nazneen@hf.co} \\
}

\begin{document}
\maketitle
\begin{abstract}
Pre-trained language models (PLMs) have been shown effective for zero-shot (\textit{0shot}) text classification. {0shot} models based on natural language inference (NLI) and next sentence prediction (NSP) employ cross-encoder architecture and infer by making a forward pass through the model for each label-text pair separately. This increases the computational cost to make inferences linearly in the number of labels. In this work, we improve the efficiency of such cross-encoder-based {0shot} models by restricting the number of likely labels using another fast \textit{base classifier}-based conformal predictor (CP) calibrated on samples labeled by the {0shot} model. Since a CP generates prediction sets with coverage guarantees, it reduces the number of target labels without excluding the most probable label based on the {0shot} model. We experiment with three intent and two topic classification datasets. With a suitable CP for each dataset, we reduce the average inference time for NLI- and NSP-based models by 25.6\% and 22.2\% respectively, without dropping performance below the predefined error rate of 1\%. 
\end{abstract}

\section{Introduction}
Zero-shot (\textit{0shot}) text classification is an important NLP problem with many real-world applications. The earliest approaches for {0shot} text classification use a similarity score between text and labels mapped to common embedding space~\citep{10.5555/1620163.1620201,10.5555/1625275.1625535,Chen2015DatalessTC,li-etal-2016-joint,Sappadla2016UsingSS,xia-etal-2018-zero}. These models calculate text and label embeddings independently and make only one forward pass over the text resulting in a minimal increase in the computation. Later approaches explicitly incorporate label information when processing the text, e.g., \citet{Yogatama2017GenerativeAD} uses generative modeling and generates text given label embedding, and \citet{rios-kavuluru-2018-shot} uses label embedding based attention over text, both requiring multiple passes over the text and increasing the computational cost.


Most recently, NLI- \cite{condoravdi-etal-2003-entailment,williams-etal-2018-broad,yin-etal-2019-benchmarking} and NSP- \cite{ma-etal-2021-issues} based {0shot} text classification formulations have been proposed. NLI and NSP make inferences by defining a representative hypothesis sentence for each label and producing a score corresponding to every pair of input text and hypothesis. To compute the score, they employ a cross-encoder architecture that is full self-attention over the text and hypothesis sentences, which requires recomputing the encoding for text and each hypothesis separately. It increases the computational cost to make inferences linearly in the number of target labels. 

NLI and NSP use large transformer-based PLMs \citep{devlin-etal-2019-bert,Liu2019RoBERTaAR,DBLP:journals/corr/abs-1910-13461} and outperform previous non-transformer-based models by a large margin. However, the size of PLMs and the number of target labels drastically reduce the prediction efficiency, increasing the computation and inference time, and may significantly increase the carbon footprint of making predictions \citep{strubell-etal-2019-energy,sustainlp-2020-sustainlp,10.1145/3381831,zhou-etal-2021-hulk}. 

In this work, we focus on the correlation between the number of labels and prediction efficiency and propose to use a conformal predictor (CP) \cite{10.5555/1062391,JMLR:v9:shafer08a} to filter out unlikely labels from the target. 
Conformal prediction provides a model-agnostic framework to generate a label set, instead of a single label prediction, within a pre-defined error rate. Consequently, we use a CP, with a small error rate we select, based on another fast \textit{base classifier} to generate candidate target labels. Candidate labels are then used with the larger NLI/NSP-based {0shot} models to make the final prediction. 

We experiment with three intent classification (SNIPS \citep{Coucke2018SnipsVP}, ATIS \citep{5700816} and HWU64 \citep{Liu2019BenchmarkingNL}) and two topic classification (AG's news and Yahoo! Answers \citep{10.5555/2969239.2969312}) datasets using {0shot} models based on a moderately sized \textit{bart-large} (NLI) \citep{lewis-etal-2020-bart} and a small \textit{bert-base} (NSP) PLM. We use four different {base} classifiers, each with different computational complexity, and a small error rate of 1\%. By using the best CP for each dataset, we reduce the average computational time by 25.6\% (22.2\%) and the average number of labels by 41.09\% (43.38\%) for NLI-(NSP-) based models.

\section{Methodology}

We improve the efficiency of NLI/NSP models by restricting the number of target labels with a Conformal Predictor (CP). Using a fast but weak {base} classifier-based CP, we produce the label set that removes some of the target classes for the {0shot} model without reducing the coverage beyond a pre-defined error rate. 

\subsection{Building a Conformal Predictor (CP) for Label Filtering}
Conformal prediction \cite{Vovk99machine-learningapplications,10.5555/1062391,JMLR:v9:shafer08a,pmlr-v128-maltoudoglou20a,https://doi.org/10.48550/arxiv.2107.07511,pmlr-v152-giovannotti21a,https://doi.org/10.48550/arxiv.2111.02592} 
generates
label sets with coverage guarantees. For a given error rate $\alpha$ and a {base} classifier $\hat{f}: x \rightarrow \mathbb{R}^K$ (here $K$ is the total number of class labels), a CP outputs a label set $\Gamma^\alpha$ that also contains true class label $y$ with probability at least $1-\alpha$. 

To build a CP, we need \textbf{calibration data} \{$(x_1,y_1),(x_2,y_2),..,(x_n,y_n)$\} and \textbf{a measure of non-conformity} $s(x_i,y_i)$ that describes the disagreement between the actual label $y_i$ and the prediction $\hat{f}(x_i)$ from the {base} classifier. 
As an example, a non-conformity score can be defined as the negative output logit of the true class. Assuming the base classifier outputs logit scores, in this case $s(x_i,y_i)$ will be $-\hat{f}(x_i)_{y_i}$.
Next, we define $\hat{q}$ to be the $\lceil(n+1)(1-\alpha)\rceil/n$ empirical quantile of scores $s(x_1,y_1),s(x_2,y_2),..,s(x_n,y_n)$ on the calibration set. Finally, for a new exchangeable test data point $x_{test}$, we output the label set $\Gamma^\alpha$ = \{$y^k:s(x_{test},y^k)<\hat{q}$\}, i.e., the classes corresponding to which the non-conformity score is lower than the $\hat{q}$. 
$\Gamma^\alpha$ is finally used with the {0shot} model to predict the final class label. 
Next, we discuss the two components of a CP, namely the calibration dataset and the non-conformity score.

\subsection{Calibration Dataset}
We require a calibration dataset that is exchangeable with the test data. However, in a typical {0shot} setting, we do not expect the availability of a human-labeled dataset. Therefore, we use the {0shot} classifier to label samples for calibration. Since our goal is to obtain a label set that contains the class label which is most probable according to the {0shot} classifier, we do not explicitly require human-labeled samples. Using model-predicted labels for calibration guarantees the required coverage.

\begin{figure*}
    \centering
    \includegraphics[width=0.323\textwidth]{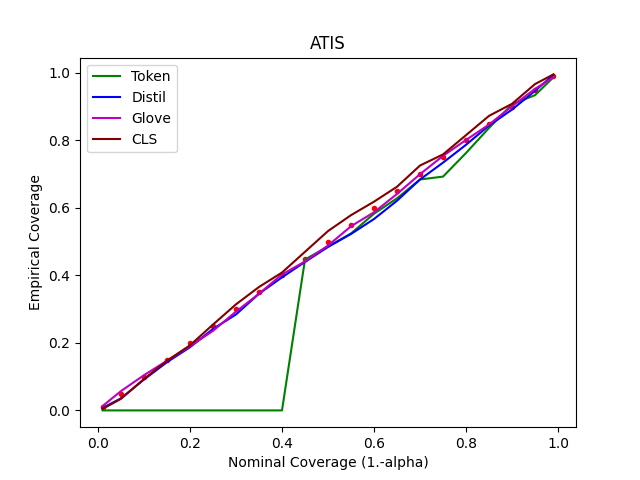}
    \includegraphics[width=0.323\textwidth]{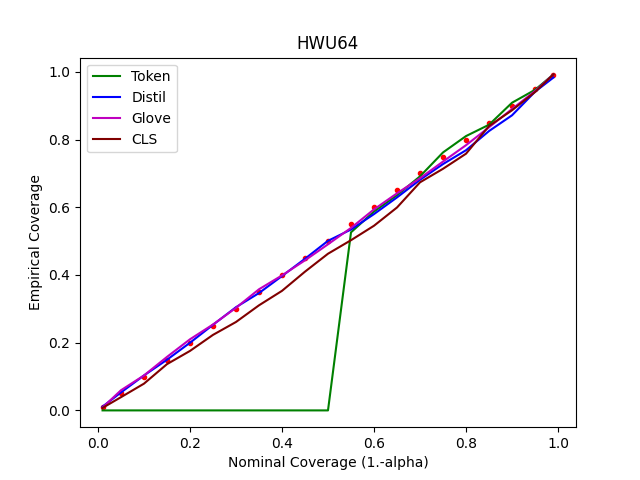}
    \includegraphics[width=0.323\textwidth]{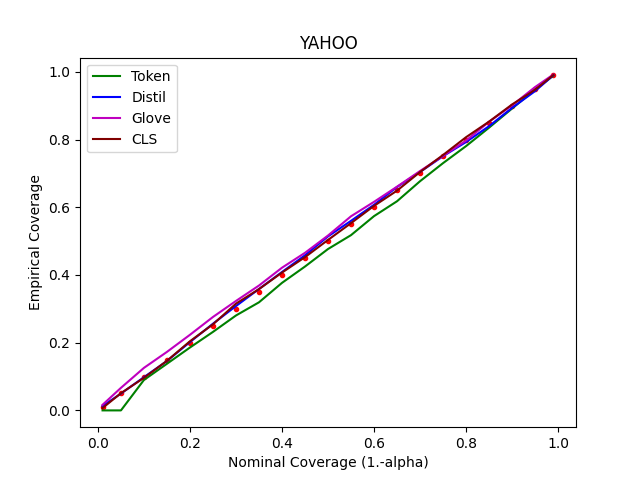}
    \caption{Coverage of CP-predicted label sets for different non-conformity score and calibration dataset on the Intent/ Topic-classification tasks with the NLI {0shot} model.}
    \label{fig:coverage}
\end{figure*}

\begin{figure*}
    \centering
    \includegraphics[width=0.323\textwidth]{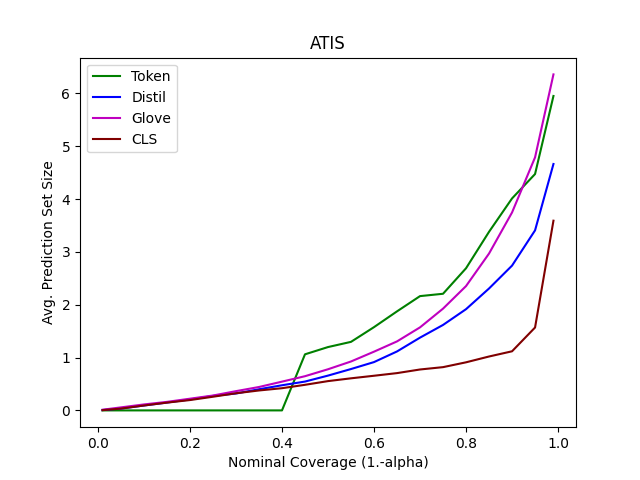}
    \includegraphics[width=0.323\textwidth]{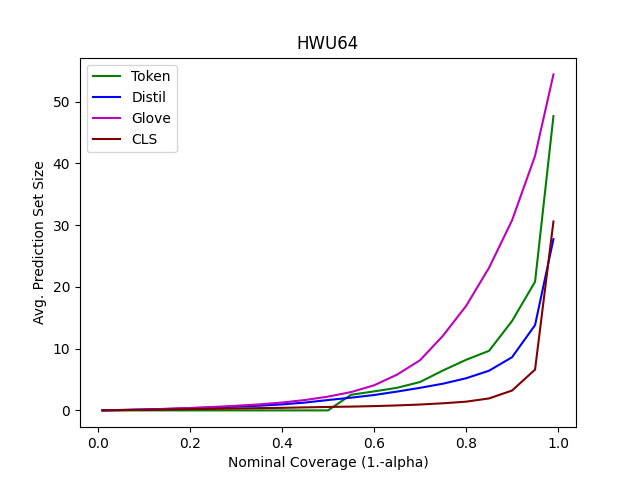}
    \includegraphics[width=0.323\textwidth]{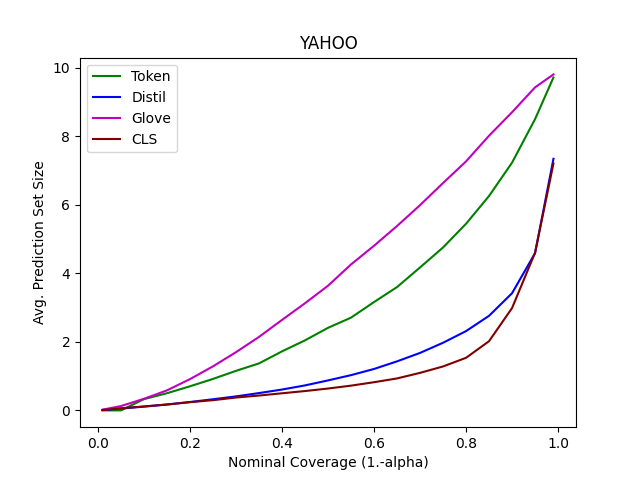}
    \caption{Average Prediction Set (APS) size for different non-conformity score and calibration dataset on the Intent/ Topic-classification tasks with the NLI {0shot} model.}
    \label{fig:aps}
\end{figure*}

\subsection{Non-Conformity score based on a {Base} Classifier} 
We want the {base} classifier to be computationally efficient when compared to the {0shot} model. 
We experiment with four {base} classifiers with different complexity for building our CPs, 
\begin{itemize}[leftmargin=*]
  \item Token Overlap (CP-Token): For each target class label ($y^k \in \{y^1,..,y^K\}$), we make a list of representative tokens ($C_w^k$) that includes all tokens in the calibration data samples corresponding to that class. Then, we define the non-conformity score using the percentage of common tokens between $C_w^k$ and the input text ($x$). Given $\#x$ defines the unique tokens in $x$, the token overlap-based non-conformity score is defined as:
  \begin{equation}
  \small
 s(x,y^k)=1.0-\frac{|C_w^k \cap x|}{\#x}
  \end{equation}
  \item Cosine Similarity (CP-Glove): Token overlap-based non-conformity score suffers from sparsity unless we use a large representative words-set for each target class label. Therefore, we also experiment with the cosine distance between the bag-of-words (BoW) representation of a target label description ($C_E^k$) and input text ($x_E$). We use static GloVe embeddings \citep{pennington2014glove} to obtain BoW representations for labels.
  \begin{equation}
  \small
 s(x,y^k)=1.-\frac{C_E^k \cdot x_E}{{\| C_E^k \|}_2 {\| x_E \|}_2}
  \end{equation}
  \item Classifier (CP-CLS): Besides the broadly applicable token overlap and cosine similarity, we propose to use a task-specific {base} classifier to generate label sets of smaller sizes.   We fine-tune a distilled \textit{bert-base} model on the data labeled using the {0shot} model and use the negative of class logits as the non-conformity scores. 
  \item Distilled NLI Model (CP-Distil): Fine-tuning a PLM (CLS) requires data labeled by the {0shot} model, which may not be always accessible. In such scenarios, we propose to use another parameter-efficient NLI {0shot} model (e.g. distil-roberta-base-nli) as the {base} classifier. While NLI-based {0shot} models are computationally expensive, they may make a good {base} classifier for relatively larger PLMs, with many parameters (e.g. GPT-3 \citep{DBLP:journals/corr/abs-2005-14165}), or when there are many target class labels (e.g. 64 labels in HWU64). We define the non-conformity score as the negative entailment probability.
\end{itemize}

\section{Experiments}
We evaluate our CP-based framework on the test set of intent (SNIPS, ATIS and HWU64) and topic (AG's news and Yahoo! Answers) classification datasets. SNIPS, ATIS, HWU64, AG's news and Yahoo! Answers have 7, 17, 64, 4 and 10 target labels, respectively. 
We use ``\textit{facebook/bart-large-nli}'' and ``\textit{bert-base-uncased}'' models from the Hugging Face hub \citep{wolf-etal-2020-transformers} as our {0shot} NLI and NSP baselines. Our experimental setup is described in appendix ($\S$\ref{experiment-apndx}).

\begin{table*}[] 
\centering
\small
\resizebox{\linewidth}{!}{
\begin{tabular}{|l|ccc|ccc|ccc|ccc|ccc|} \hline
 & \multicolumn{3}{c|}{ATIS} & \multicolumn{3}{c|}{HWU64} & \multicolumn{3}{c|}{SNIPS} & \multicolumn{3}{c|}{AGNEWS}  & \multicolumn{3}{c|}{YAHOO}\\ \hline
 & A $\uparrow$ & T $\downarrow$ & |L| $\downarrow$ & A $\uparrow$ & T $\downarrow$ & |L| $\downarrow$ & A $\uparrow$ & T $\downarrow$ & |L| $\downarrow$ & A $\uparrow$ & T $\downarrow$ & |L| $\downarrow$ & A $\uparrow$ & T $\downarrow$ & |L| $\downarrow$ \\ \hline
 \multicolumn{16}{|c|}{Entailment: \textit{bart-large-mnli}} \\ \hline
Full & \textbf{71.5} & 1.0 & 17.0 & 49.3 & 1.0 & 64.0 & 76.7 & 1.0 & 7.0 & {75.9} & 1.0 & 4.0 & 51.8 & 1.0  & 10.0  \\
CP-Token & 70.1 & \textbf{0.63} & \textbf{9.1} & 49.4 & 0.79 & 47.6 &  75.8 & \textbf{0.88}  & 5.9  & 75.5 & \textbf{0.91} & 3.5 & 51.4 & 0.91 &  9.7\\
CP-Glove &  71.0 & \underline{1.01} & 16.7  & 49.3 & 1.0 & 63.4 & 76.0 & 0.98 & 6.8 & 75.0 & 0.99  & 3.9 & 51.4 & \underline{1.05}  & 9.8 \\
CP-CLS  & \textbf{71.5} & 0.98 & 9.2 & 49.3 & 0.71 & 30.59 & \textbf{77.0} & 0.98 & \textbf{3.5} & \textbf{76.2} & 0.99 & \textbf{2.4} & 51.7 & \textbf{0.80} & 8.1\\
CP-Distil & \textbf{71.5} & 0.98  & 14.0 & \textbf{49.7} & \textbf{0.50}  & \textbf{27.7} & 76.7 & 0.91  & {4.6} & 75.9 & 0.94  & {2.9} & \textbf{51.9} & {0.83}  & \textbf{7.2} \\ \hline 
 \multicolumn{16}{|c|}{Next Sentence Prediction: \textit{bert-base-uncased}} \\ \hline
Full & 23.6 & 1.0 & 17.0 & 47.2 & 1.0 & 64.0 & 81.7 & 1.0 & 7.0 & 79.3 & 1.0 & 4.0 & 52.0 & 1.0 & 10.0 \\
CP-Token & 23.1 & \textbf{0.41} & \textbf{5.3} & 47.2 & 0.85 & 53.8 & 81.0 & 0.94 & 6.0 & 78.9 & \textbf{0.96} & 3.5 & 52.0 & 0.97 & 9.6 \\
CP-Glove & \textbf{23.9} & 0.99 & 16.4 & 47.1 & 1.0 & 63.0 & 80.8 & 1.0 & 6.8 & 78.3 & 0.99 & 3.9 & 51.8 & 1.0 & 9.6 \\
CP-CLS  & 23.6 & 0.61 & 6.1 & \textbf{47.3} & \textbf{0.69} & \textbf{39.3} & 81.8 & \textbf{0.88} & 4.4 & \textbf{80.4} & \underline{1.41} & \textbf{2.1} & 52.3 & \textbf{0.95} & \textbf{7.8} \\
CP-Distil & 23.7 & \underline{1.51} & 16.1 & \textbf{47.3} & \underline{1.15} & 57.7 & \textbf{82.7} & \underline{1.97} & \textbf{4.2} & 80.1 & \underline{2.13} & 3.0 & \textbf{52.6} & \underline{1.69} & 8.8 \\ \hline 
\end{tabular}
}
\caption{Performance comparison of different {base} classifiers. We use the error rate $\alpha$ of 0.01, and report accuracy (A), average time to make inference (T) and average prediction set size (|L|). We underline the cases where a CP increases the inference time. \label{table:result}}
\end{table*}

In Fig. \ref{fig:coverage} and \ref{fig:aps}, we plot the empirical coverage and the average prediction set (APS) size of four {base} classifiers on ATIS, HWU64 and Yahoo! Answers datasets.
\textit{Coverage} defines the proportion of samples for which the predicted set contains the {0shot} model-predicted label. \textit{APS size} equals the average number of labels in the set predicted by the CP {base} classifier. 

In Table \ref{table:result}, we compare the accuracy (A), average inference time (T) and the APS size (|L|) used with the {0shot} model. Average inference time is reported relative to the full model. ``{Full}'' represents the {0shot} model that uses the full label set.  During inference, we create one batch for all text-label pairs in a sample. For instance, with 64 labels, a batch includes 64 text-label pairs where each pair consists of text and one label. This allows us to measure the reduction in inference time while fully utilizing the available compute resources. 

\subsection{Results}
\noindent \textbf{A CP achieves a valid coverage}. 
We find that for smaller values of $\alpha$, all four {base} classifiers achieve valid coverage (Fig. \ref{fig:coverage}), i.e., empirical and nominal coverages are identical, implying that we can use a CP to filter unlikely target labels without dropping the performance below a low predefined error rate $\alpha$. 
For larger $\alpha$s (>$\sim$0.5), empirical coverage drops to 0 on intent datasets for token overlap-based non-conformity score. The reduced coverage for token overlap at lower $\alpha$ results from an empty label set, as evident from 0 APS size in Fig. \ref{fig:aps}. 

\noindent \textbf{A CP reduces the average number of labels for the {0shot} model.} 
We find that a
stronger {base} classifier (CLS and Distil) provides lower APS size for the same empirical (or nominal) coverage (Fig. \ref{fig:aps}). On average, CP-CLS provides the lowest APS size, reducing the average number of labels for both {0shot} models by roughly 41\% (Table \ref{table:result}). This suggests that fine-tuning a {base} classifier should be preferred when unlabelled samples are easily available.

\noindent \textbf{A simpler and efficient CP {base} classifier may reduce the inference time the most.}
We observe that CP-Token achieves the best inference time with the NLI model on ATIS, SNIPS and AG's news datasets, and with the NSP model on ATIS and AG's news datasets. On the other hand, it achieves the lowest APS size for both models only on the ATIS dataset. Minimal complexity for calculating token-overlap adds negligible overhead to the {0shot} model, thus, achieving the best speed up despite higher APS size in several cases. 


\noindent \textbf{A CP {base} classifier needs to be computationally inexpensive.}
CP-Distil improves inference time for the NLI model on all datasets but fails to do so for the NSP model, despite reduced APS size. This ineffectiveness is explained by the comparable inference time for the {base} (distil-nli) classifier and the {0shot} NSP model. When building a CP, it is imperative to select a {base} classifier that is computationally economical relative to the {0shot} model.

\noindent \textbf{A CP improves efficiency the most on the dataset with many labels.}
We observe the maximum speed up on HWU64  and ATIS datasets. This is unsurprising given the relatively higher number of possible target labels for both datasets, emphasizing the benefit of a CP for tasks with many target labels. 

\noindent \textbf{A CP performs comparably to the {0shot} model.}
CP-based label filtering retains the performance of the corresponding models that use a full label set. Among the four {base} classifiers, CP-Token performs the worst (-0.46\% absolute drop) and CP-Distil performs the best (+0.31\% absolute gain) on the average accuracy. It is noteworthy that the accuracy increases in many cases, suggesting that pruning label space using a CP may remove noisy labels and boost the performance.

\subsection{Applying a CP in Practical Applications}
Our experiments show that the inference speed-up from a CP depends on the sizes of the zero-shot model, the base classifier, and the label space. A strong base classifier (e.g., CP-CLS, CP-Distil) often gives better APS size leading to faster zero-shot inference. But it is also slow in generating the label set for the 0shot model. On the other hand, weaker base classifiers are fast but generate larger prediction sets resulting in slower 0shot inference. Given the trade-off, a stronger base classifier model such as BERT/ RoBERTa (or distilled model) makes a better choice when the label space is large (e.g., 64 for HWU64) and (or) the zero-shot model is large (e.g., \textit{bart-large-mnli}). Otherwise, a faster base classifier (e.g., token overlap matching) would be ideal.



\section{Discussion and Future Work}
In this work, we show that CP-based target label filtering improves the efficiency of NLI- and NSP-based zero-shot text classification models. Our CP framework is generalizable to many formulations of zero- or few-shot models. For instance, prompt-based models that auto-regressively generate verbalizers \citep{schick-schutze-2021-just} are very slow, as the number of forward passes increases with the number of labels and verbalizer’s tokens \citep{karimi-mahabadi-etal-2022-prompt}. Our conformal prediction framework can be directly used there to reduce the number of labels and improve efficiency. Additionally, the proposed conformal prediction framework can be used to filter training examples when constructing prompts for 
\textit{in-context learning}  \citep{DBLP:journals/corr/abs-2005-14165}, where we prime PLM with a sequence of training examples. For a task with many target labels, a naively constructed prompt would require at least one training example for each label. However, with CP, we can limit the number of labels (and consequently the number of training examples) in the prompt, minimizing the computational cost. 


In the future, we  will explore newer methods to build conformal predictors that can further reduce the average prediction set size and inference time, as well as boost the performance of a {0shot} model.

\section{Limitations}
The datasets utilized in this research contain texts in English and thus mainly represent the culture of the English-speaking populace. Gender or age biases may also exist in the datasets, and pre-trained models may also exhibit these biases. In addition, though the likelihood is slim, additional biases may also be introduced by CP-based label filtering.

We recommend that any subsequent usage of the proposed technique clearly states that the conformal predictor was used for label filtering. In addition, while results suggest that our framework is generally effective for different zero-shot tasks, it should not be utilized on any new task without thorough evaluation, including evaluating for ethical or social risks.

\section*{Acknowledgement}
We would like to thank the anonymous reviewers for their feedback.

\bibliography{anthology,custom}

\begin{thebibliography}{37}
\expandafter\ifx\csname natexlab\endcsname\relax\def\natexlab#1{#1}\fi

\bibitem[{Angelopoulos and
  Bates(2021)}]{https://doi.org/10.48550/arxiv.2107.07511}
Anastasios~N. Angelopoulos and Stephen Bates. 2021.
\newblock \href {https://doi.org/10.48550/ARXIV.2107.07511} {A gentle
  introduction to conformal prediction and distribution-free uncertainty
  quantification}.

\bibitem[{Brown et~al.(2020)Brown, Mann, Ryder, Subbiah, Kaplan, Dhariwal,
  Neelakantan, Shyam, Sastry, Askell, Agarwal, Herbert{-}Voss, Krueger,
  Henighan, Child, Ramesh, Ziegler, Wu, Winter, Hesse, Chen, Sigler, Litwin,
  Gray, Chess, Clark, Berner, McCandlish, Radford, Sutskever, and
  Amodei}]{DBLP:journals/corr/abs-2005-14165}
Tom~B. Brown, Benjamin Mann, Nick Ryder, Melanie Subbiah, Jared Kaplan,
  Prafulla Dhariwal, Arvind Neelakantan, Pranav Shyam, Girish Sastry, Amanda
  Askell, Sandhini Agarwal, Ariel Herbert{-}Voss, Gretchen Krueger, Tom
  Henighan, Rewon Child, Aditya Ramesh, Daniel~M. Ziegler, Jeffrey Wu, Clemens
  Winter, Christopher Hesse, Mark Chen, Eric Sigler, Mateusz Litwin, Scott
  Gray, Benjamin Chess, Jack Clark, Christopher Berner, Sam McCandlish, Alec
  Radford, Ilya Sutskever, and Dario Amodei. 2020.
\newblock \href {http://arxiv.org/abs/2005.14165} {Language models are few-shot
  learners}.
\newblock \emph{CoRR}, abs/2005.14165.

\bibitem[{Chang et~al.(2008)Chang, Ratinov, Roth, and
  Srikumar}]{10.5555/1620163.1620201}
Ming-Wei Chang, Lev Ratinov, Dan Roth, and Vivek Srikumar. 2008.
\newblock Importance of semantic representation: Dataless classification.
\newblock In \emph{Proceedings of the 23rd National Conference on Artificial
  Intelligence - Volume 2}, AAAI'08, page 830–835. AAAI Press.

\bibitem[{Chen et~al.(2015)Chen, Xia, Jin, and Carroll}]{Chen2015DatalessTC}
Xingyuan Chen, Yunqing Xia, Peng Jin, and John~A. Carroll. 2015.
\newblock Dataless text classification with descriptive lda.
\newblock In \emph{AAAI}.

\bibitem[{Condoravdi et~al.(2003)Condoravdi, Crouch, de~Paiva, Stolle, and
  Bobrow}]{condoravdi-etal-2003-entailment}
Cleo Condoravdi, Dick Crouch, Valeria de~Paiva, Reinhard Stolle, and Daniel~G.
  Bobrow. 2003.
\newblock \href {https://aclanthology.org/W03-0906} {Entailment, intensionality
  and text understanding}.
\newblock In \emph{Proceedings of the {HLT}-{NAACL} 2003 Workshop on Text
  Meaning}, pages 38--45.

\bibitem[{Coucke et~al.(2018)Coucke, Saade, Ball, Bluche, Caulier, Leroy,
  Doumouro, Gisselbrecht, Caltagirone, Lavril, Primet, and
  Dureau}]{Coucke2018SnipsVP}
Alice Coucke, Alaa Saade, Adrien Ball, Th{\'e}odore Bluche, Alexandre Caulier,
  David Leroy, Cl{\'e}ment Doumouro, Thibault Gisselbrecht, Francesco
  Caltagirone, Thibaut Lavril, Ma{\"e}l Primet, and Joseph Dureau. 2018.
\newblock Snips voice platform: an embedded spoken language understanding
  system for private-by-design voice interfaces.
\newblock \emph{ArXiv}, abs/1805.10190.

\bibitem[{Devlin et~al.(2019)Devlin, Chang, Lee, and
  Toutanova}]{devlin-etal-2019-bert}
Jacob Devlin, Ming-Wei Chang, Kenton Lee, and Kristina Toutanova. 2019.
\newblock \href {https://doi.org/10.18653/v1/N19-1423} {{BERT}: Pre-training of
  deep bidirectional transformers for language understanding}.
\newblock In \emph{Proceedings of the 2019 Conference of the North {A}merican
  Chapter of the Association for Computational Linguistics: Human Language
  Technologies, Volume 1 (Long and Short Papers)}, pages 4171--4186,
  Minneapolis, Minnesota. Association for Computational Linguistics.

\bibitem[{Dey et~al.(2021)Dey, Ding, Ferrell, Kapper, Lovig, Planchon, and
  Williams}]{https://doi.org/10.48550/arxiv.2111.02592}
Neil Dey, Jing Ding, Jack Ferrell, Carolina Kapper, Maxwell Lovig, Emiliano
  Planchon, and Jonathan~P Williams. 2021.
\newblock \href {https://doi.org/10.48550/ARXIV.2111.02592} {Conformal
  prediction for text infilling and part-of-speech prediction}.

\bibitem[{Gabrilovich and Markovitch(2007)}]{10.5555/1625275.1625535}
Evgeniy Gabrilovich and Shaul Markovitch. 2007.
\newblock Computing semantic relatedness using wikipedia-based explicit
  semantic analysis.
\newblock In \emph{Proceedings of the 20th International Joint Conference on
  Artifical Intelligence}, IJCAI'07, page 1606–1611, San Francisco, CA, USA.
  Morgan Kaufmann Publishers Inc.

\bibitem[{Giovannotti and Gammerman(2021)}]{pmlr-v152-giovannotti21a}
Patrizio Giovannotti and Alex Gammerman. 2021.
\newblock \href {https://proceedings.mlr.press/v152/giovannotti21a.html}
  {Transformer-based conformal predictors for paraphrase detection}.
\newblock In \emph{Proceedings of the Tenth Symposium on Conformal and
  Probabilistic Prediction and Applications}, volume 152 of \emph{Proceedings
  of Machine Learning Research}, pages 243--265. PMLR.

\bibitem[{Karimi~Mahabadi et~al.(2022)Karimi~Mahabadi, Zettlemoyer, Henderson,
  Mathias, Saeidi, Stoyanov, and Yazdani}]{karimi-mahabadi-etal-2022-prompt}
Rabeeh Karimi~Mahabadi, Luke Zettlemoyer, James Henderson, Lambert Mathias,
  Marzieh Saeidi, Veselin Stoyanov, and Majid Yazdani. 2022.
\newblock \href {https://doi.org/10.18653/v1/2022.acl-long.254} {Prompt-free
  and efficient few-shot learning with language models}.
\newblock In \emph{Proceedings of the 60th Annual Meeting of the Association
  for Computational Linguistics (Volume 1: Long Papers)}, pages 3638--3652,
  Dublin, Ireland. Association for Computational Linguistics.

\bibitem[{Lewis et~al.(2019)Lewis, Liu, Goyal, Ghazvininejad, Mohamed, Levy,
  Stoyanov, and Zettlemoyer}]{DBLP:journals/corr/abs-1910-13461}
Mike Lewis, Yinhan Liu, Naman Goyal, Marjan Ghazvininejad, Abdelrahman Mohamed,
  Omer Levy, Veselin Stoyanov, and Luke Zettlemoyer. 2019.
\newblock \href {http://arxiv.org/abs/1910.13461} {{BART:} denoising
  sequence-to-sequence pre-training for natural language generation,
  translation, and comprehension}.
\newblock \emph{CoRR}, abs/1910.13461.

\bibitem[{Lewis et~al.(2020)Lewis, Liu, Goyal, Ghazvininejad, Mohamed, Levy,
  Stoyanov, and Zettlemoyer}]{lewis-etal-2020-bart}
Mike Lewis, Yinhan Liu, Naman Goyal, Marjan Ghazvininejad, Abdelrahman Mohamed,
  Omer Levy, Veselin Stoyanov, and Luke Zettlemoyer. 2020.
\newblock \href {https://doi.org/10.18653/v1/2020.acl-main.703} {{BART}:
  Denoising sequence-to-sequence pre-training for natural language generation,
  translation, and comprehension}.
\newblock In \emph{Proceedings of the 58th Annual Meeting of the Association
  for Computational Linguistics}, pages 7871--7880, Online. Association for
  Computational Linguistics.

\bibitem[{Li et~al.(2016)Li, Zheng, Tian, Hu, Iyer, and
  Sycara}]{li-etal-2016-joint}
Yuezhang Li, Ronghuo Zheng, Tian Tian, Zhiting Hu, Rahul Iyer, and Katia
  Sycara. 2016.
\newblock \href {https://aclanthology.org/C16-1252} {Joint embedding of
  hierarchical categories and entities for concept categorization and dataless
  classification}.
\newblock In \emph{Proceedings of {COLING} 2016, the 26th International
  Conference on Computational Linguistics: Technical Papers}, pages 2678--2688,
  Osaka, Japan. The COLING 2016 Organizing Committee.

\bibitem[{Liu et~al.(2019{\natexlab{a}})Liu, Eshghi, Swietojanski, and
  Rieser}]{Liu2019BenchmarkingNL}
Xingkun Liu, Arash Eshghi, Pawel Swietojanski, and Verena Rieser.
  2019{\natexlab{a}}.
\newblock Benchmarking natural language understanding services for building
  conversational agents.
\newblock In \emph{IWSDS}.

\bibitem[{Liu et~al.(2019{\natexlab{b}})Liu, Ott, Goyal, Du, Joshi, Chen, Levy,
  Lewis, Zettlemoyer, and Stoyanov}]{Liu2019RoBERTaAR}
Yinhan Liu, Myle Ott, Naman Goyal, Jingfei Du, Mandar Joshi, Danqi Chen, Omer
  Levy, Mike Lewis, Luke Zettlemoyer, and Veselin Stoyanov. 2019{\natexlab{b}}.
\newblock Roberta: A robustly optimized bert pretraining approach.
\newblock \emph{ArXiv}, abs/1907.11692.

\bibitem[{Loshchilov and Hutter(2019)}]{loshchilov2018decoupled}
Ilya Loshchilov and Frank Hutter. 2019.
\newblock \href {https://openreview.net/forum?id=Bkg6RiCqY7} {Decoupled weight
  decay regularization}.
\newblock In \emph{International Conference on Learning Representations}.

\bibitem[{Ma et~al.(2021)Ma, Yao, Lin, and Zhao}]{ma-etal-2021-issues}
Tingting Ma, Jin-Ge Yao, Chin-Yew Lin, and Tiejun Zhao. 2021.
\newblock \href {https://doi.org/10.18653/v1/2021.acl-short.99} {Issues with
  entailment-based zero-shot text classification}.
\newblock In \emph{Proceedings of the 59th Annual Meeting of the Association
  for Computational Linguistics and the 11th International Joint Conference on
  Natural Language Processing (Volume 2: Short Papers)}, pages 786--796,
  Online. Association for Computational Linguistics.

\bibitem[{Maltoudoglou et~al.(2020)Maltoudoglou, Paisios, and
  Papadopoulos}]{pmlr-v128-maltoudoglou20a}
Lysimachos Maltoudoglou, Andreas Paisios, and Harris Papadopoulos. 2020.
\newblock \href {https://proceedings.mlr.press/v128/maltoudoglou20a.html}
  {Bert-based conformal predictor for sentiment analysis}.
\newblock In \emph{Proceedings of the Ninth Symposium on Conformal and
  Probabilistic Prediction and Applications}, volume 128 of \emph{Proceedings
  of Machine Learning Research}, pages 269--284. PMLR.

\bibitem[{Moosavi et~al.(2020)Moosavi, Fan, Shwartz, Glava{\v{s}}, Joty, Wang,
  and Wolf}]{sustainlp-2020-sustainlp}
Nafise~Sadat Moosavi, Angela Fan, Vered Shwartz, Goran Glava{\v{s}}, Shafiq
  Joty, Alex Wang, and Thomas Wolf, editors. 2020.
\newblock \href {https://aclanthology.org/2020.sustainlp-1.0}
  {\emph{Proceedings of SustaiNLP: Workshop on Simple and Efficient Natural
  Language Processing}}. Association for Computational Linguistics, Online.

\bibitem[{Pennington et~al.(2014)Pennington, Socher, and
  Manning}]{pennington2014glove}
Jeffrey Pennington, Richard Socher, and Christopher~D. Manning. 2014.
\newblock \href {http://www.aclweb.org/anthology/D14-1162} {Glove: Global
  vectors for word representation}.
\newblock In \emph{Empirical Methods in Natural Language Processing (EMNLP)},
  pages 1532--1543.

\bibitem[{Rios and Kavuluru(2018)}]{rios-kavuluru-2018-shot}
Anthony Rios and Ramakanth Kavuluru. 2018.
\newblock \href {https://doi.org/10.18653/v1/D18-1352} {Few-shot and zero-shot
  multi-label learning for structured label spaces}.
\newblock In \emph{Proceedings of the 2018 Conference on Empirical Methods in
  Natural Language Processing}, pages 3132--3142, Brussels, Belgium.
  Association for Computational Linguistics.

\bibitem[{Sappadla et~al.(2016)Sappadla, Nam, Menc{\'i}a, and
  F{\"u}rnkranz}]{Sappadla2016UsingSS}
Prateek~Veeranna Sappadla, Jinseok Nam, Eneldo~Loza Menc{\'i}a, and Johannes
  F{\"u}rnkranz. 2016.
\newblock Using semantic similarity for multi-label zero-shot classification of
  text documents.
\newblock In \emph{ESANN}.

\bibitem[{Schick and Sch{\"u}tze(2021)}]{schick-schutze-2021-just}
Timo Schick and Hinrich Sch{\"u}tze. 2021.
\newblock \href {https://doi.org/10.18653/v1/2021.naacl-main.185} {It{'}s not
  just size that matters: Small language models are also few-shot learners}.
\newblock In \emph{Proceedings of the 2021 Conference of the North American
  Chapter of the Association for Computational Linguistics: Human Language
  Technologies}, pages 2339--2352, Online. Association for Computational
  Linguistics.

\bibitem[{Schwartz et~al.(2020)Schwartz, Dodge, Smith, and
  Etzioni}]{10.1145/3381831}
Roy Schwartz, Jesse Dodge, Noah~A. Smith, and Oren Etzioni. 2020.
\newblock \href {https://doi.org/10.1145/3381831} {Green ai}.
\newblock \emph{Commun. ACM}, 63(12):54–63.

\bibitem[{Shafer and Vovk(2008)}]{JMLR:v9:shafer08a}
Glenn Shafer and Vladimir Vovk. 2008.
\newblock \href {http://jmlr.org/papers/v9/shafer08a.html} {A tutorial on
  conformal prediction}.
\newblock \emph{Journal of Machine Learning Research}, 9(12):371--421.

\bibitem[{Strubell et~al.(2019)Strubell, Ganesh, and
  McCallum}]{strubell-etal-2019-energy}
Emma Strubell, Ananya Ganesh, and Andrew McCallum. 2019.
\newblock \href {https://doi.org/10.18653/v1/P19-1355} {Energy and policy
  considerations for deep learning in {NLP}}.
\newblock In \emph{Proceedings of the 57th Annual Meeting of the Association
  for Computational Linguistics}, pages 3645--3650, Florence, Italy.
  Association for Computational Linguistics.

\bibitem[{Tur et~al.(2010)Tur, Hakkani-Tür, and Heck}]{5700816}
Gokhan Tur, Dilek Hakkani-Tür, and Larry Heck. 2010.
\newblock \href {https://doi.org/10.1109/SLT.2010.5700816} {What is left to be
  understood in atis?}
\newblock In \emph{2010 IEEE Spoken Language Technology Workshop}, pages
  19--24.

\bibitem[{Vovk et~al.(2005)Vovk, Gammerman, and Shafer}]{10.5555/1062391}
Vladimir Vovk, Alex Gammerman, and Glenn Shafer. 2005.
\newblock \emph{Algorithmic Learning in a Random World}.
\newblock Springer-Verlag, Berlin, Heidelberg.

\bibitem[{Vovk et~al.(1999)Vovk, Gammerman, and
  Saunders}]{Vovk99machine-learningapplications}
Volodya Vovk, Alex Gammerman, and Craig Saunders. 1999.
\newblock Machine-learning applications of algorithmic randomness.
\newblock In \emph{In Proceedings of the Sixteenth International Conference on
  Machine Learning}, pages 444--453. Morgan Kaufmann.

\bibitem[{Williams et~al.(2018)Williams, Nangia, and
  Bowman}]{williams-etal-2018-broad}
Adina Williams, Nikita Nangia, and Samuel Bowman. 2018.
\newblock \href {https://doi.org/10.18653/v1/N18-1101} {A broad-coverage
  challenge corpus for sentence understanding through inference}.
\newblock In \emph{Proceedings of the 2018 Conference of the North {A}merican
  Chapter of the Association for Computational Linguistics: Human Language
  Technologies, Volume 1 (Long Papers)}, pages 1112--1122, New Orleans,
  Louisiana. Association for Computational Linguistics.

\bibitem[{Wolf et~al.(2020)Wolf, Debut, Sanh, Chaumond, Delangue, Moi, Cistac,
  Rault, Louf, Funtowicz, Davison, Shleifer, von Platen, Ma, Jernite, Plu, Xu,
  Scao, Gugger, Drame, Lhoest, and Rush}]{wolf-etal-2020-transformers}
Thomas Wolf, Lysandre Debut, Victor Sanh, Julien Chaumond, Clement Delangue,
  Anthony Moi, Pierric Cistac, Tim Rault, Rémi Louf, Morgan Funtowicz, Joe
  Davison, Sam Shleifer, Patrick von Platen, Clara Ma, Yacine Jernite, Julien
  Plu, Canwen Xu, Teven~Le Scao, Sylvain Gugger, Mariama Drame, Quentin Lhoest,
  and Alexander~M. Rush. 2020.
\newblock \href {https://www.aclweb.org/anthology/2020.emnlp-demos.6}
  {Transformers: State-of-the-art natural language processing}.
\newblock In \emph{Proceedings of the 2020 Conference on Empirical Methods in
  Natural Language Processing: System Demonstrations}, pages 38--45, Online.
  Association for Computational Linguistics.

\bibitem[{Xia et~al.(2018)Xia, Zhang, Yan, Chang, and Yu}]{xia-etal-2018-zero}
Congying Xia, Chenwei Zhang, Xiaohui Yan, Yi~Chang, and Philip Yu. 2018.
\newblock \href {https://doi.org/10.18653/v1/D18-1348} {Zero-shot user intent
  detection via capsule neural networks}.
\newblock In \emph{Proceedings of the 2018 Conference on Empirical Methods in
  Natural Language Processing}, pages 3090--3099, Brussels, Belgium.
  Association for Computational Linguistics.

\bibitem[{Yin et~al.(2019)Yin, Hay, and Roth}]{yin-etal-2019-benchmarking}
Wenpeng Yin, Jamaal Hay, and Dan Roth. 2019.
\newblock \href {https://doi.org/10.18653/v1/D19-1404} {Benchmarking zero-shot
  text classification: Datasets, evaluation and entailment approach}.
\newblock In \emph{Proceedings of the 2019 Conference on Empirical Methods in
  Natural Language Processing and the 9th International Joint Conference on
  Natural Language Processing (EMNLP-IJCNLP)}, pages 3914--3923, Hong Kong,
  China. Association for Computational Linguistics.

\bibitem[{Yogatama et~al.(2017)Yogatama, Dyer, Ling, and
  Blunsom}]{Yogatama2017GenerativeAD}
Dani Yogatama, Chris Dyer, Wang Ling, and Phil Blunsom. 2017.
\newblock Generative and discriminative text classification with recurrent
  neural networks.
\newblock \emph{ArXiv}, abs/1703.01898.

\bibitem[{Zhang et~al.(2015)Zhang, Zhao, and LeCun}]{10.5555/2969239.2969312}
Xiang Zhang, Junbo Zhao, and Yann LeCun. 2015.
\newblock Character-level convolutional networks for text classification.
\newblock In \emph{Proceedings of the 28th International Conference on Neural
  Information Processing Systems - Volume 1}, NIPS'15, page 649–657,
  Cambridge, MA, USA. MIT Press.

\bibitem[{Zhou et~al.(2021)Zhou, Chen, Jin, and Wang}]{zhou-etal-2021-hulk}
Xiyou Zhou, Zhiyu Chen, Xiaoyong Jin, and William~Yang Wang. 2021.
\newblock \href {https://doi.org/10.18653/v1/2021.eacl-demos.39} {{HULK}: An
  energy efficiency benchmark platform for responsible natural language
  processing}.
\newblock In \emph{Proceedings of the 16th Conference of the European Chapter
  of the Association for Computational Linguistics: System Demonstrations},
  pages 329--336, Online. Association for Computational Linguistics.

\end{thebibliography}
\bibliographystyle{acl_natbib}

\appendix \label{sec:appendix}

\section{Experimental Details} \label{experiment-apndx}
We use the entire training set of intent datasets and 5000 samples from the validation set of topic datasets to calibrate CP-Token, CP-Glove and CP-Distil. For CP-CLS, we use the entire training set of intent datasets and 2500 samples from the validation set of topic datasets to train the {base} classifier, and the entire validation set of intent datasets and 2500 samples from the validation set of topic datasets for calibration. 

For the CP-Distil {base} classifier, we use ``\textit{cross-encoder/nli-distilroberta-base}'' model from the Hugging Face hub. We describe the procedure for building CP-CLS in $\S$\ref{base-cls-appendix}. Note that we only require the text to be classified (without labels) for calibrating {base} classifiers and training CP-CLS {base} classifier. We use the {0shot} model to label the corresponding training and calibration samples.



\subsection{CP-CLS: Training and Evaluation} \label{base-cls-appendix}
We fine-tune \textit{distilbert-base} PLM for 15 epochs with the batch size of 16. We use AdamW optimizer \citep{loshchilov2018decoupled} with default hyper-parameters, warm-up steps of 100 and weight decay of 0.01. We choose the checkpoint with the highest accuracy on the calibration dataset. The accuracy of the CP-CLS base classifier with respect to the labels generated by the {0shot} model is reported in Table \ref{cp-cls-acc}.

\begin{table}[h] 
\centering
\small
\resizebox{\columnwidth}{!}{
\begin{tabular}{|l|ccc|cc|} \hline
 & ATIS & HWU64 & SNIPS & AGNEWS & YAHOO \\ \hline
NLI & 84.99 & 67.28 & 88.0 & 66.88 & 87.64 \\
NSP & 82.80 & 71.18 & 89.40 & 73.16 & 88.24 \\\hline
\end{tabular}}
\caption{Accuracy of the CP-CLS base classifier with respect to the labels generated by the {0shot} model on the calibration dataset. \label{cp-cls-acc}}
\end{table}

\section{Calibration dataset}

\begin{figure}[!h]
     \centering
     \includegraphics[width=0.44\textwidth]{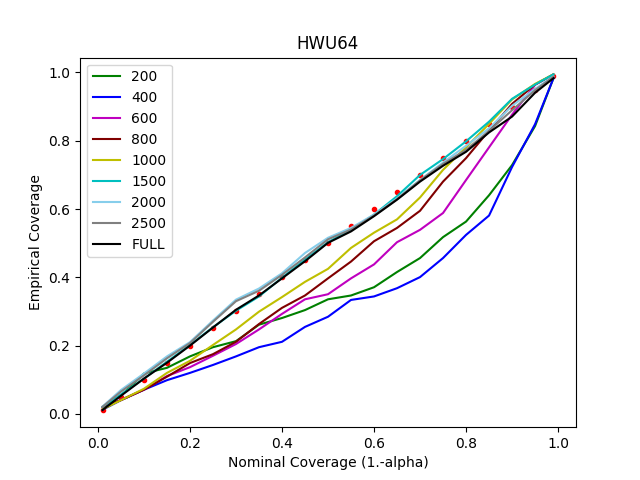}
     \caption{Coverage of CP-predicted label sets for different size of calibration set.  \label{calibration-size} 
     }
\end{figure}

\begin{figure}[!h]
     \centering
     \includegraphics[width=0.44\textwidth]{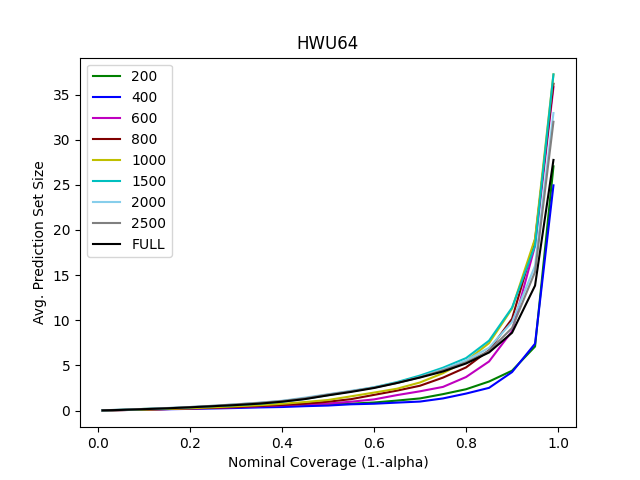}
     \caption{APS size with different size of calibration set.  \label{calibration-size-aps}
     }
\end{figure}

\begin{figure}[!h]
     \centering
     \includegraphics[width=0.44\textwidth]{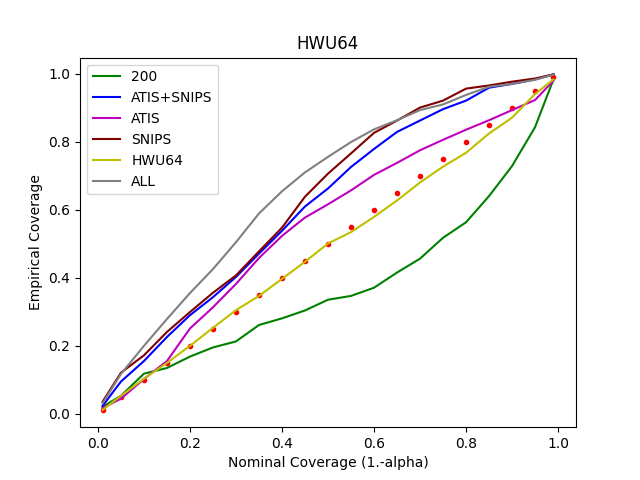}
     \caption{Coverage of CP-predicted label sets for calibration set from another dataset.   \label{transfer}
     }
\end{figure}

\begin{figure}[!h]
     \centering
     \includegraphics[width=0.44\textwidth]{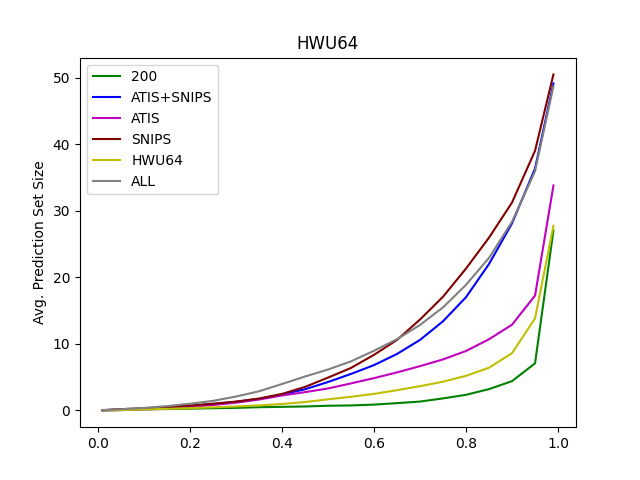}
     \caption{APS size for calibration set from another dataset. \label{transfer-size}
     }
\end{figure}

In our experiments, we assumed the availability of text from the target task for calibration (and training a {base} classifier). While we can use a {0shot} base classifier (e.g., CP-Token /Glove/ Distil), we still require a few samples for the CP calibration. Next, we analyze the number of samples required to build a CP and the transferability of a CP from another different dataset in Fig. \ref{calibration-size}, \ref{calibration-size-aps}, \ref{transfer} and \ref{transfer-size}. These plots belong to the NLI-based {0shot} model on the HWU64 dataset with the corresponding best-performing CP-Distil non-conformity score. 200 (up to 2500) in Fig. \ref{calibration-size} and \ref{calibration-size-aps} represents the number of samples used for calibration. The full model is calibrated on the entire dataset. In Fig. \ref{transfer} and \ref{transfer-size}, SNIPS+ATIS is calibrated using the samples from SNIPS and ATIS datasets, and ALL is calibrated using the samples from SNIPS, ATIS, Yahoo! Answers and AG's news datasets. 
We describe our findings below.

\noindent \textbf{A small sized-calibration set with low chosen $\alpha$ improves {0shot} classification efficiency without dropping the performance.}
We find that the empirical coverage is worse than the nominal coverage for smaller-sized calibration data (200-1000) (Fig. \ref{calibration-size}). A model using calibration data with 1500 or more samples has identical empirical and nominal coverages. At the same time, the difference in empirical and nominal coverage is negligible for all calibration data sizes at a low error rate ($\alpha=0.01$), as evident from the same starting point for all plots (at $1-\alpha=0.99$). Consequently, we can use the conformal prediction framework even with a small calibration set provided we use a low value for $\alpha$. Next, as shown in Fig. \ref{calibration-size-aps}, the dependence between the APS size and the size of calibration data is not linear. CP-Distil models based on 200, 400 and full calibration samples obtain comparable APS sizes. Additionally, APS sizes for other models are significantly better than the full label set (maximum APS size of ~37 for models calibrated on 1000 and 1500 samples vs full label set size of 64).

\noindent \textbf{We can use a calibration set from another dataset only if the $\alpha$ is set to a low value}
We observe that none of the ATIS, SNIPS, ATIS+SNIPS or ALL calibration sets obtains the same empirical and nominal coverage. But,
similar to our observations on the size of the calibration set, samples from another dataset can be used for calibration (Fig. \ref{transfer}), provided we use a low $\alpha$. 
However, the resulting APS size is larger than that of the model that uses calibration data from the target task (Fig. \ref{transfer-size}).


\end{document}